\documentclass[lettersize,journal]{IEEEtran}
\usepackage{amsmath,amsfonts}
\usepackage{algorithmic}
\usepackage{algorithm}
\usepackage{array}
\usepackage[caption=false,font=normalsize,labelfont=sf,textfont=sf]{subfig}
\usepackage{textcomp}
\usepackage{stfloats}
\usepackage{url}
\usepackage{verbatim}
\usepackage{graphicx}
\usepackage{cite}
\usepackage{multirow}
\usepackage{adjustbox}
\usepackage{booktabs}
\def\eg{\emph{e.g.}}

\def\ie{\emph{i.e.}}

\usepackage{pifont} 

\begin{document}


\title{AdapSNE: Adaptive Fireworks-Optimized and Entropy-Guided Dataset Sampling for Edge DNN Training}

\author{Boran Zhao$^{1}$, Hetian Liu$^{1}$, Zihang Yuan$^{1}$, Li Zhu$^{1}$, Fan Yang$^{2}$, Lina Xie$^{3}$ Tian Xia$^{4}$, Wenzhe Zhao$^{4}$ \\ and Pengju Ren$^{4}$~\IEEEmembership{Member,~IEEE}



\thanks{- $^{1}$Boran Zhao, Hetian Liu, Zihang Yuan and Li Zhu are with the School of Software Engineering.}

\thanks{- $^{2}$Fan Yang is the School of Computer Science and Technology.}

\thanks{- $^{3}$Lina Xie is with the the First Affiliated Hospital of Xi'an Jiaotong University.}

\thanks{- $^{4}$All other authors are with the National Key Laboratory of Human-Machine Hybrid Augmented Intelligence, National Engineering Research Center for Visual Information and Applications, and Institute of Artificial Intelligence and Robotics, Xi'an Jiaotong University, Xi’an, Shaanxi, China.} 

\thanks{- E-mail: pengjuren@xjtu.edu.cn (Corresponding Author).}
}

\markboth{Journal of \LaTeX\ Class Files,~Vol.~14, No.~8, August~2021}
{Shell \MakeLowercase{\textit{et al.}}: A Sample Article Using IEEEtran.cls for IEEE Journals}


\maketitle

\begin{abstract}
Training deep neural networks (DNNs) directly on edge devices has attracted increasing attention, as it offers promising solutions to challenges such as domain adaptation and privacy preservation. However, conventional DNN training typically requires large-scale datasets, which imposes prohibitive overhead on edge devices—particularly for emerging large language model (LLM) tasks. To address this challenge, a DNN-free method (\ie, dataset sampling without DNN), named NMS (Near-Memory Sampling), has been introduced. By first conducting dimensionality reduction of the dataset and then performing exemplar sampling in the reduced space, NMS avoids the architectural bias inherent in DNN-based methods and thus achieves better generalization. However, The state-of-the-art, NMS, suffers from two limitations: (1) The mismatch between the search method and the non-monotonic property of the perplexity error function leads to the emergence of outliers in the reduced representation; (2) Key parameter (\ie, target perplexity) is selected empirically, introducing arbitrariness and leading to uneven sampling. These two issues lead to \textit{representative bias} of examplars, resulting in degraded accuracy. To address these issues, we propose AdapSNE, which integrates an efficient non-monotonic search method—namely, the Fireworks Algorithm (FWA)—to suppress outliers, and employs entropy-guided optimization to enforce uniform sampling, thereby ensuring representative training samples and consequently boosting training accuracy. 
To cut the edge-side cost arising from the iterative computations of FWA search and entropy-guided optimization, we design an accelerator with custom dataflow and time—multiplexing markedly reducing on-device training energy and area. Experimental results show that our AdapSNE outperforms the SOTA DNN-based (\ie, DQAS) and DNN-free (\ie, NMS) methods. On small-scale image datasets, AdapSNE achieves improvements of 4.4\% and 0.85\%, respectively. On large-scale image datasets, the gains increase to 8.3\% over DQAS and 2.5\% over NMS. Furthermore, on the MMLU LLM benchmark, AdapSNE achieves average performance improvements of 3.5\% over DQAS and 2.4\% over NMS.  
\end{abstract}



\section{Introduction}
\IEEEPARstart{D}{eep} Neural Networks have achieved remarkable success across a wide range of domains \cite{he2016deep,he2017mask,dosovitskiy2020image,ma2018shufflenet,xia2022energy,zhao2022remap,zhao2023hipu}. However, one of the reasons for the accuracy improvement in DNNs is their reliance on training models with massive amounts of data, which incurs significant overhead. For instance, training the Gemini\cite{team2023gemini} large language model required approximately $\mathcal{O}(10^3)$ MWh of energy, enough to power 94 households for an entire year\cite{nms}. In addition, when a model is deployed in a new domain, it typically requires retraining or fine-tuning on data specific to the target scenario. For example, applying a DNN model trained on ImageNet\cite{deng2009imagenet} to medical X-ray image analysis (\eg, pneumonia detection) typically requires retraining with a medical dataset (\eg, CheXpert\cite{chexpert}). With the widespread adoption of the Internet of Things and autonomous driving, there is an increasing need to deploy a large number of edge devices across diverse application scenarios. However, due to communication latency, data privacy concerns, and the heterogeneity of data across devices, training models in the cloud becomes impractical. As a result, there is an increasing need to retrain or fine-tune neural networks directly on edge devices. 

However, the limited computational resources and battery capacities of edge devices make it difficult to support on-device model training. For example, a typical edge device—such as an electric vehicle—has a battery capacity of only $10^{-1}$ MWh, which is several orders of magnitude lower than the energy required to train large language models (LLMs). To address the challenge of efficient DNNs training on edge devices, researchers have proposed a series of dataset compression methods, aiming to reduce training overhead by eliminating redundancy within the training data.

\begin{figure}[!t]
\centering
\includegraphics[width=1\linewidth]{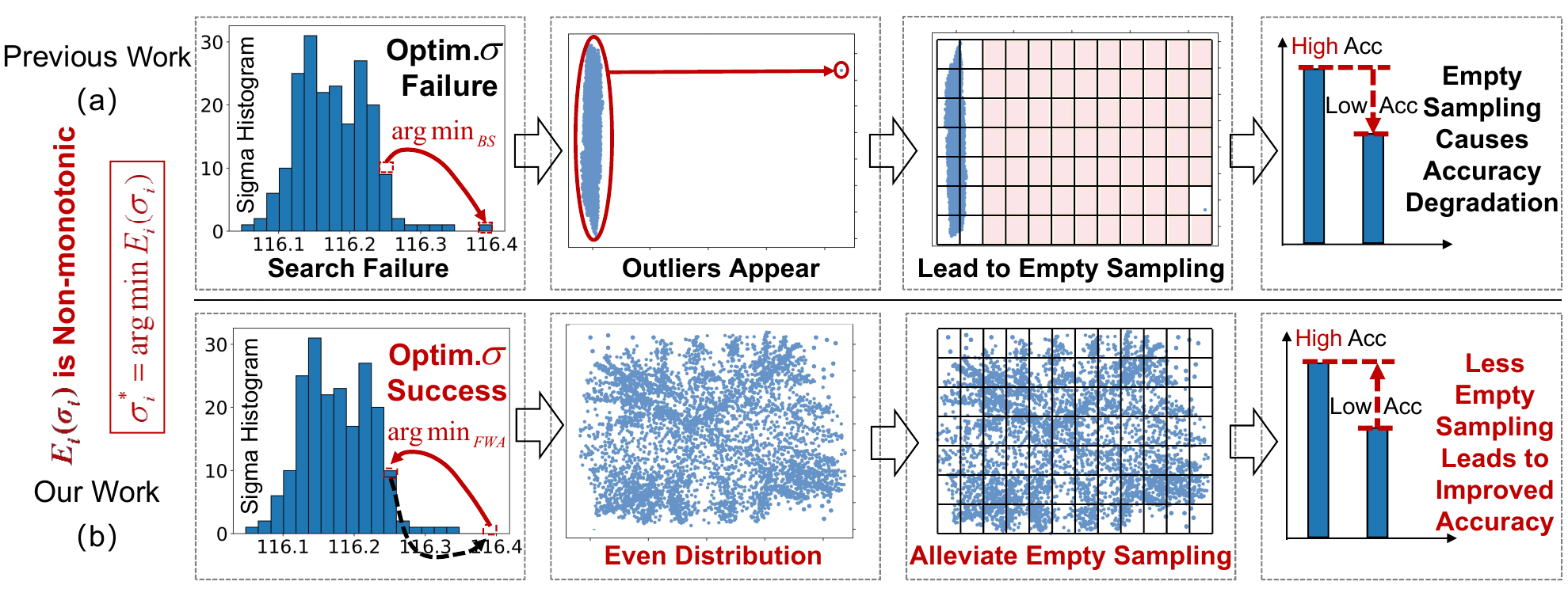}
\caption{Traditional methods for optimal search often lead to the emergence of outliers\cite{nms}. }
\label{outlier}
\end{figure}


Dataset compression methods can be broadly categorized into two types: data synthesis and data selection. Data synthesis \cite{loo2022efficient,cui2023scaling,zhao2023improved,cazenavette2023generalizing,zhao2023dataset} typically treats the samples to be sampled as learnable parameters and introduces additional gradient descent steps into the standard training process to synthesize the new representative samples. However, due to the additional gradient descent, these methods incur a significantly high time overhead (\eg, \cite{wang2018dataset,deng2022remember,zhao2020dataset}). To address this issue, researchers have proposed data selection based dataset compression algorithms\cite{killamsetty2021glister}, which evaluate pairwise distances between samples using specific metrics and extract a small number of representative samples within local neighborhoods to construct the condensed training set. This strategy reduces the compression overhead of the additional gradient descent caused by data synthesis\cite{sener2017active,chen2012super}. Data selection can be further categorized into DNN-based\cite{toneva2018empirical,margatina2021active} and DNN-free~\cite{nms} approaches. DNN-based methods rely on extracting sample features from the inference results of a DNN, followed by examplar selection based on these features. However, this process inherently introduces architectural bias, as the extracted features are tightly coupled with the specific architecture and parameters of the DNN backbone, ultimately leading to poor generalization (\eg, Using images synthesized by ResNet18~\cite{DQ}, the training accuracy on CNext drops from 84.1\% to 52.8\%, a decrease of 31.3\%).

\begin{figure}[!t]
    \centering
    \includegraphics[width=0.85\linewidth]{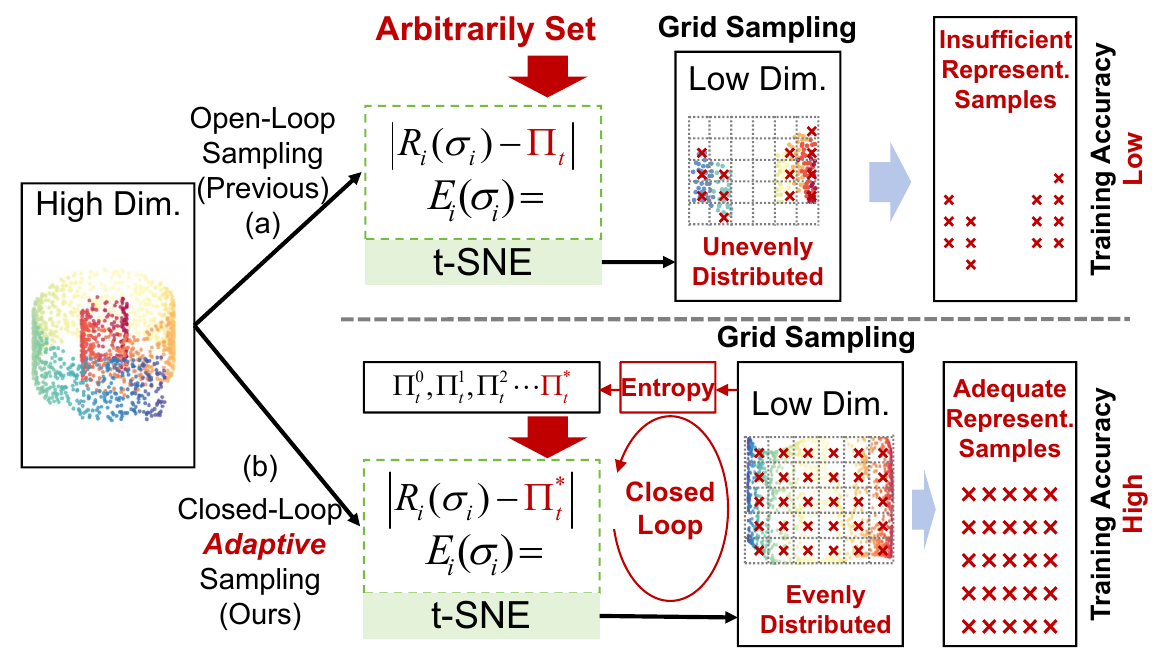}
    \caption{The previous SOTA method, NMS\cite{nms}, adopts an open-loop sampling strategy, whereas our proposed method employs a closed-loop adaptive sampling approach. }
    \label{open_vs_close}
\end{figure}

\begin{figure}[!t]
\centering
\includegraphics[width=0.75\linewidth]{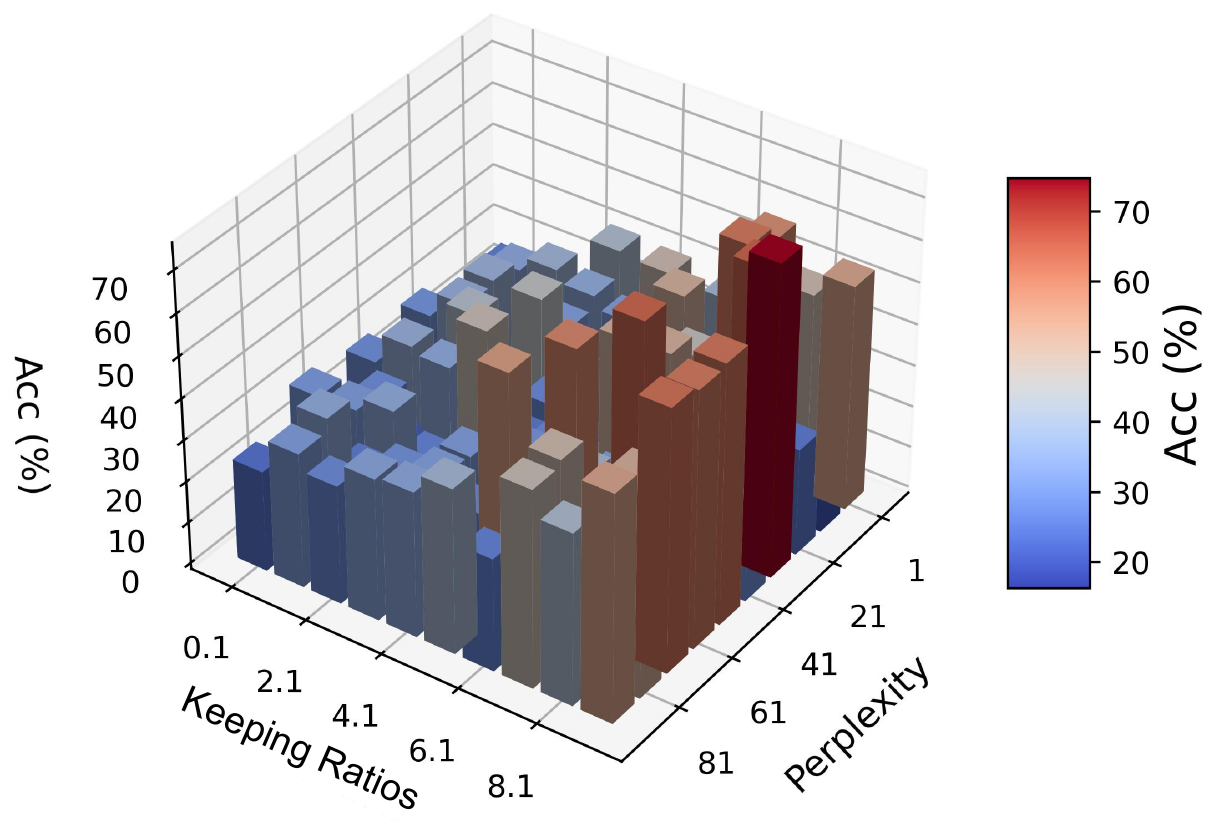}
\caption{The optimal perplexity for achieving the best training accuracy varies at different compression rates. Data is from training ResNet-18 on CIFAR-10.}
\label{opt_perpl}
\end{figure}

To address this issue, researchers have proposed DNN-free methods, which do not rely on DNN to compute the feature distances between samples. As a result, they avoid architectural bias and typically exhibit better generalization, which is why we focus on this class of methods in our work. The SOTA DNN-free method, NMS, utilizes t-SNE to evaluate the distances between samples and perform sampling. By avoiding the involvement of DNNs during sampling, it improves the generalization of the selected samples to some extent. However, our research finds that this method, which uses open-loop sampling, inevitably suffers from two issues: 

1) Due to the mismatch between the search method and the non-monotonic property of the perplexity error function, outliers appear in the reduced space, as shown in Fig.~\ref{outlier} (a). 2) Key parameter (\ie, target perplexity) is selected empirically, as shown in Fig.~\ref{open_vs_close} (a). The arbitrariness of this selection, due to the fact that the optimal perplexity for different scenarios corresponds to different training accuracies, as shown in Fig.~\ref{opt_perpl}, leads to uneven distribution of the condensed samples.

These two issues lead to insufficient representative samples (\ie, \textit{representative bias} problem) selected by NMS, ultimately resulting in poor training accuracy. Therefore, efficiently obtaining a compressed and effective training dataset on edge devices remains an open problem.

To address this, we propose AdapSNE, a novel adaptive fireworks-optimized and entropy-guided dataset sampling framework. 
To overcome outliers problem, we leverage an efficient Fireworks Algorithm (FWA) to explore the perplexity error function. To mitigate the challenge of \textit{representative bias}, an entropy-guided optimization automatically tunes the target perplexity. Finally, a custom-dataflow, time-multiplexed accelerator curbs the iterative overhead of FWA and entropy-guided optimization, markedly cutting on-device training energy and silicon area.


Experimental results show that AdapSNE outperforms the SOTA DNN-based (\ie, DQAS) and DNN-free (\ie, NMS) methods across multiple datasets. On small-scale image datasets, AdapSNE achieves improvements of 4.4\% over DQAS and 0.85\% over NMS. On large-scale image datasets, it improves by 8.3\% and 2.5\%, respectively. On the MMLU large language model benchmark, AdapSNE shows an average improvement of 3.5\% over DQAS and 2.4\% over NMS.

Our main contributions can be summarized as follows:
\begin{itemize}
\item{We provide a rigorous mathematical proof, for the first time, demonstrating that the perplexity error function is non-monotonic, thereby theoretically uncovering the underlying mathematical principles that explain the infeasibility of traditional search methods (\eg, binary search).}




\item {We present a novel fireworks-based algorithm that efficiently explores the non-monotonic perplexity error function, thereby mitigating the accuracy degradation caused by outliers during dimensionality reduction.}

\item{We propose an entropy-guided optimization method that overcomes \textit{representative bias}, ensures even sampling, and thus improves edge-training accuracy}
\item {We propose an accelerator with custom dataflow and time—multiplexing markedly reducing the overhead of edge device training. }
\end{itemize}

\section{{Background and Motivation}}
Dataset compression methods can be categorized into two types: data synthesis and dataset selection\cite{survey}. The goal of both methods is to construct a smaller dataset that can replace the original large dataset for training. Data synthesis methods generate new samples that are not present in the original dataset. In contrast, dataset selection methods do not create new data but instead select representative samples based on specific metrics. We provide a detailed overview of both approaches in the following sections.

\subsection{Data Synthesis}
Data synthesis methods, also known as dataset distillation, were first formally introduced by Wang \textit{et al.}~\cite{wang2018dataset}. Subsequently, a series of dataset distillation methods have been proposed. Meta-model matching~\cite{deng2022remember} generates synthetic data by matching the prediction errors between the synthetic dataset and the original dataset. More recently, several studies have explored generating new samples by matching the byproducts produced during the training process of synthetic and original datasets. Depending on the type of byproduct being matched, these methods can be categorized into parameter matching~\cite{zhao2023dataset}, gradient matching \cite{zhao2020dataset,kim2022dataset}, and trajectory matching methods \cite{cui2023scaling,liu2023dream}. However, these methods generate synthetic samples by introducing numerous gradient backpropagation operations into the standard training process, which results in substantial computational and energy overhead, making them impractical for deployment in edge scenarios.

\subsection{Dataset Selection}
To address the high computational overhead in data synthesis methods, researchers have proposed dataset selection methods, which can be classified into DNN-based and DNN-free approaches.

DNN-based methods utilize DNNs to extract features from samples for dataset compression. For example, geometric methods \cite{sener2017active,chen2012super}, loss-based methods \cite{toneva2018empirical}, and decision boundary-based methods \cite{margatina2021active}. It is worth noting that all DNN-based methods rely on the forward inference results of the DNN as features to select representative samples. However, this process inevitably incorporates the architecture and weight parameters of the feature-extracting DNN, introducing architectural bias in the compressed dataset. This leads to poor generalization on other types of DNNs, resulting in accuracy degradation. For example,  the images synthesized by the DQ \cite{DQ} method for ResNet-18 result in a significant accuracy drop of 25.7\% (from 84.1\% to 58.4\%) on ViT and 31.3\% (from 84.1\% to 52.8\%) on CNext. 
Moreover, the DNN computation requires enormous resources, leading to significant overhead. For instance, the time spent on extracting representative images using DQAS is 45 times that of the DNN training time~\cite{DQAS}. To address this issue, researchers have proposed DNN-free methods.

DNN-free methods typically estimate feature distances between samples in high-dimensional space using Gaussian kernels. They then project the samples into a low-dimensional space while preserving their relative distances, and perform sample selection in the reduced space. For instance, the current SOTA DNN-free method, NMS, employs t-SNE with Gaussian kernels to estimate sample pairwise distances and performs grid sampling in the reduced space. By avoiding the utilization of DNNs during sample selection process, it eliminates architectural bias and thereby improving the model's generalization.

\subsection{Challenges and Opportunities}

However, the NMS method is subject to two unavoidable challenges.

\begin{itemize}
\item{Non-monotonicity of the perplexity error function: Traditional t-SNE determines the local perplexity using a binary search strategy, under the assumption that the perplexity error function is monotonic. However, through rigorous mathematical proof, we show that the perplexity error function is not monotonic. This non-monotonicity leads to failures in locating the function’s optimal value, resulting in outliers in the low-dimensional projection, which further causes degraded accuracy. Although NMS claims to alleviate this issue by employing a differential evolution algorithm\cite{de} to search the perplexity error function, our experimental results show that due to the function’s non-monotonicity, the instability remains unresolved. }


\item{Target perplexity is arbitrary: NMS selects the target perplexity in t-SNE based on empirical heuristics, which introduces considerable arbitrariness and makes it difficult to adapt to specific dataset compression tasks. However, our experiments show that for typical dataset compression scenarios, such as training ResNet-18 on CIFAR-10, different keeping ratios (KR) require different perplexity values to achieve optimal accuracy, as illustrated in Fig.~\ref{opt_perpl}.}

\end{itemize}


Due to the above two challenges, the NMS method can introduce \textit{representative bias} of the examplars, leading to degraded training accuracy. To address this, we propose an fireworks-optimized and entropy-guided dataset sampling framework, namely AdapSNE. 

\section{Proposed Algorithm}


\subsection{Preliminary of t-SNE}
\subsubsection{Probability Distribution in the High-Dimension}
Given a dataset $\mathbf{X} = \{x_1, x_2, \dots, x_N\}$ containing $N$ samples, each sample is associated with a Gaussian distribution centered at $x_i$, with standard deviation $\sigma_i$. Accordingly, the conditional probability of similarity between $x_i$ and $x_j$ is defined as:
\begin{equation}
    p_{j|i} = \frac{\exp\left(-\frac{\|x_i - x_j\|^2}{2\sigma_i^2} \right)}{\sum_{k \ne i} \exp\left(-\frac{\|x_i - x_k\|^2}{2\sigma_i^2} \right)}
\end{equation}

The joint similarity between $x_i$ and $x_j$ is:
\begin{equation}
    p_{ij}=\frac{p_{i|j}+p_{j|i}}{2N}
\end{equation}

The standard deviation $\sigma_i$ is determined by the target perplexity $\Pi_t$ in the perplexity error function, and the target perplexity is a manually specified target value based on empirical heuristics. This is one of the key factors contributing to the instability of t-SNE dimensionality reduction:
\begin{equation}
    \sigma_i^* = \arg\min_{\sigma_i} \left| R_i(\sigma_i) - \Pi_t \right|    
\end{equation}

Traditional t-SNE determines the standard deviation via binary search, while NMS employs a differential evolution algorithm for standard deviation optimization. The local perplexity $R_i$ is calculated as follows:
\begin{equation}
    R_i (\sigma_i) = 2^{ - \sum_j p_{j|i} \log_2 p_{j|i} }
\end{equation}

In the following section, we prove that the perplexity error function is inherently non-monotonic. As a result, binary search cannot guarantee convergence, which constitutes the second major cause of t-SNE's instability.  

\subsubsection{Probability Distribution in the Low-Dimension}

Given a set $\mathbf{Y}=\{y_1, y_2, \dots, y_N\}$ representing the low-dimensional embeddings of the high-dimensional dataset $\mathbf{X} = \{x_1, x_2, \dots, x_N\}$. Assuming that the data points $y_i$ follow a student's t-distribution with one degree of freedom, the joint probability $q_{ij}$ between two points $y_i$ and $y_j$ in the low-dimensional space is defined as follows:

\begin{equation}
    q_{ij} = \frac{\left(1 + \|y_i - y_j\|^2\right)^{-1}}{\sum_{k \ne l} \left(1 + \|y_k - y_l\|^2\right)^{-1}}
\end{equation}

\subsubsection{Computing the Coordinates in the Low-Dimensional Space}
The similarity between the probability distribution $P$ in the high-dimensional space and $Q$ in the low-dimensional space is measured using Kullback-Leibler (KL) divergence.
\begin{equation}
    C = KL(P \| Q) = \sum_{i} \sum_{j} p_{ij} \log \left( \frac{p_{ij}}{q_{ij}} \right)
\end{equation}


\subsection{Proving the Non-Monotonicity of the Perplexity Error Function}
Based on the assumptions made earlier, we will prove the non-monotonicity of the perplexity error function which is determined by the local perplexity $R_i(\sigma_i)$ in this section. For convenience, we define $ H(p_{j|i}) = - \sum_{j \ne i} p_{j|i} \log_2p_{j|i} $, so the original local perplexity $R_i(\sigma_i)$ becomes:
\begin{equation}
    R_i = 2^{H(p_{j|i})}
\end{equation}

Now, the derivative of the function $R_i$ with respect to $\sigma_i$ is:
\begin{equation}
    \frac{dR_i}{d\sigma_i} = R_i \cdot \ln(2) \cdot \frac{dH(p_{j|i})}{d\sigma_i}
\end{equation}

To compute the derivative of $H$ with respect to $\sigma_i$ , we have
\begin{equation}
    \frac{dH}{d\sigma_i} = - \sum_{j \ne i} \left( \frac{dp_{j|i}}{d\sigma_i} \log_2p_{j|i} + p_{j|i} \cdot \frac{1}{p_{j|i}} \frac{dp_{j|i}}{d\sigma_i} \cdot \frac{1}{\ln 2} \right)
\end{equation} 

For convenience, we define:
\begin{equation}
    D_i = \sum_{k \ne i} p_{k|i} \|x_i - x_k\|^2 
\end{equation}

Now, the derivative of $p_{j|i}$  with respect to $\sigma_i$  is:
\begin{equation}
    \frac{dp_{j|i}}{d\sigma_i} = \frac{p_{j|i}}{\sigma_i^3} \left( \|x_i - x_j\|^2 - D_i \right)
\end{equation}

Thus, we obtain the following:
\begin{equation}
    \frac{dH}{d\sigma_i} = - \frac{1}{\sigma_i^3 \ln 2} \sum_{j \ne i}  p_{j|i} \left( \|x_i - x_j\|^2 - D_i \right) \left( \ln p_{j|i} + 1 \right) 
\end{equation}

To determine the sign of $\frac{dH}{d\sigma_i}$, we analyze the sign of each factor in the summation:
\begin{enumerate}
    \item {First Factor: $p_{j|i}$ \\ $p_{j|i}$ is a probability and always positive:
    \[
    p_{j|i} > 0
    \]
    }
    \item {Second Factor: $\|x_i-x_j\|^2-D_i$ \\ $D_i$ is the expected squared distance from $x_i$ to other points $x_k$, weighted by $p_{k|i}$. Therefore, the term $\|x_i-x_j\|^2 -D_i$ can be positive, zero, or negative. 

    }
    \item {Third Factor: $\ln{p_{j|i}}+1$ \\ Since $0 < p_{j|i} <1$, we know that $\ln{p_{j|i}} < 0$. Thus, the term $\ln{p_{j|i}}+1$ can be either positive or negative, depending on the value of $p_{j|i}$. 
    }
\end{enumerate}

Based on the above analysis, it can be concluded that the perplexity error function is non-monotonic. Notably, this is the first work to rigorously prove the non-monotonicity of the function from a mathematical perspective. However, traditional methods typically assume its monotonicity to achieve fast convergence, such as the binary search in t-SNE and the differential evolution algorithm in NMS. As discussed earlier, this incorrect assumption leads to the emergence of outliers during dimensionality reduction and results in degraded training accuracy in dataset compression methods like NMS. In addition, the selection of the target perplexity is arbitrary, which is the key reason for the uneven distribution in the low-dimensional space. To address these issues, we propose a fireworks-optimized and entropy-guided dataset sampling framework.

\subsection{AdapSNE Overview}



\begin{algorithm}[!t]
\caption{Framework of AdapSNE}
\label{alg_ov}
\begin{algorithmic}
\STATE \textbf{Input:} Dataset $\mathbf{X} = \{x_1, x_2, \dots, x_N\}$, an original target perplexity $\Pi_t^0$
\STATE \textbf{Output:} A set of representative data $Examplar = \{e_1, e_2, \dots, e_m\}$


\vspace{0.5em}

\STATE 1: Given $\mathbf{X}$, define the objective $f(\sigma_i)=\lvert R_i(\sigma_i)-\Pi_t\rvert$. Then obtain the optimal $\sigma_i^*$ via FWA-optimized search and use it in t-SNE to compute the low-dimensional coordinate $\mathbf{Y}$

\STATE 2: Given $\mathbf{Y}$, calculate the entropy for the Entropy-Guided optimization, with the first iteration serving as the baseline $H_b$
\STATE 3: Calculate the theoretical maximum entropy $H_{\text{max}}$ for the given grid number $g \times g$:
\[
H_{\text{max}}=log(g^2)
\]
\STATE 4: Set entropy threshold $H_0 = (H_{\text{max}} - H_b) \times \alpha + H_b$

\vspace{0.3em}
\STATE $k=0$
\WHILE{$H < H_0$}
    \IF{$k==0$}
    \STATE $\Pi_t^{k+1} = \Pi_t^{k} + \lceil \Delta \Pi_t \rceil$
  
    \ELSE 
    \STATE Calculate $\Delta H/ \Delta \Pi_t$ with small $\Delta \Pi_t$     
    \[
    \Delta H/ \Delta \Pi_t = \frac{H(\Pi_t^{k} + \Delta \Pi_t) - H(\Pi_t^{k})}{\Delta \Pi_t}
    \]
    \[
    \Pi_t^{k+1} = \Pi_t^{k} + \lceil \frac{H(\Pi_t^k) - H(\Pi_t^{k-1})}{\Delta H/ \Delta \Pi_t + \epsilon} \rceil
    \] 
    \STATE where \( \epsilon \) is a small constant used to avoid division by zero
    \ENDIF
    
    \STATE Redo steps 1, obtain $\mathbf{Y'}$ 
    \STATE Recompute the entropy $H$ based on $\mathbf{Y'}$
    \STATE $k=k+1$
\ENDWHILE

\vspace{0.5em}
\STATE Grid sampling, obtain $Examplar = \{e_1, e_2, \dots, e_m\}$
\STATE \textbf{Return:} $Examplar$

\end{algorithmic}
\end{algorithm}

\begin{figure}[!t]
\centering
\includegraphics[width=0.48\linewidth]{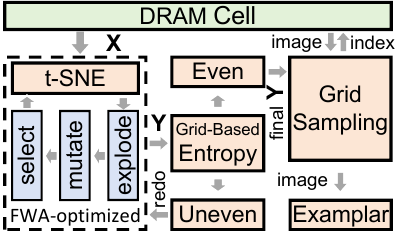}
\caption{The overview of AdapSNE accelerator. }
\label{adapSNE_overview}
\end{figure}

As shown in Fig.~\ref{adapSNE_overview}, the AdapSNE circuit consists of three parts: the FWA-optimized t-SNE module, the entropy calculation module, and the grid sampling module. Firstly, the FWA-optimized t-SNE module reads the dataset $\mathbf{X}$ from the DRAM cell and computes the low-dimensional coordinates $\mathbf{Y}$. During this stage, the perplexity error function is searched via the FWA, which consists of three phases: explosion, mutation, and selection. Then the entropy calculation module computes the entropy of $\mathbf{Y}$. 
If the distribution is even (\ie, the entropy exceeds a threshold $H_0$), grid sampling is performed and the representative samples are output. If the distribution is uneven, the target perplexity $\Pi_t$ is adaptively adjusted, the new coordinates $\mathbf{Y'}$ are recomputed, and the distribution is estimated until the low-dimensional coordinates distribute evenly. 

The detail is described in Algorithm~\ref{alg_ov}.
Similar to the circuit description, the algorithm first reduces the dimensionality of $\mathbf{X}$ to obtain coordinate $\mathbf{Y}$, and then evaluates whether the distribution is even to decide whether to redo the process or perform grid sampling. The difference is that Algorithm~\ref{alg_ov} specifies in detail the threshold $H_0$ and the method to update the target perplexity $\Pi_t$. The threshold $H_0$ is set to $H_0 = (H_{\text{max}} - H_b) \times \alpha + H_b$. We set $\alpha$ to 80\% based on~\cite{entropy}, which indicates that the distribution is sufficiently even. 
For the update of $\Pi_t$, we use Newton's method to compute the ratio of the change in entropy to the change in perplexity within a small range $\Delta \Pi_t$ to determine the update direction of $\Pi_t$. It is worth noting that, following\cite{newton}, we set a parameter $\epsilon$ to prevent division by zero. 
 

\subsection{Proposed Fireworks-Based Optimization and Acceleration Circuit}

\subsubsection{Challenges in Optimizing the Perplexity Error Function}
The perplexity error function induces an astronomically large search space, thereby complicating the optimization.  
Consider a dataset of \(200\) samples, where the standard deviation of each sample can take any integer value in \([1,100]\); the resulting search space comprises \(100^{200}\) candidate solutions. As discussed earlier, this space is non-monotonic, which further hampers convergence. Consequently, conventional algorithms, such as the binary search in t-SNE\cite{tsne_hiton} and the differential evolution in NMS\cite{nms}, are ill-suited for efficiently traversing such a vast and irregular landscape.

\subsubsection{Proposed Method}To address this challenge, we adopt a fireworks-based algorithm which allows simultaneous random variations of 200 variables in a single iteration. This capability enables rapid adjustment of multiple variables, making it well-suited for this optimization problem. The fireworks algorithm (FWA) is a nature-inspired optimization algorithm \cite{tan2010fireworks}. It imitates the explosion process of fireworks to solve large-space and non-monotonic problem. 

The algorithm is based on the following principles. 
\ding{182} Explosion: Candidate solutions, called fireworks, are generated in the search space. Each firework produces sparks, whose number and range are determined by the fitness of the firework. Better solutions generate more sparks in a smaller range, promoting local search, while poorer solutions produce fewer sparks in a larger range to explore globally.
\ding{183} Mutation: To increase diversity and avoid premature convergence, random sparks are generated using a special mutation to explore new areas of the search space.
\ding{184} Selection: A subset of the generated sparks and the original fireworks are selected to form the next generation. This selection balances exploration and exploitation. FWA has been successfully applied in diverse fields, including engineering optimization, machine learning, and operations research.

\begin{algorithm}[!t]
\caption{FWA-Optimized Search}
\label{alg_fwa}
\begin{algorithmic}
\STATE \textbf{Input:} Objective function $f(\sigma_i) = |R_i(\sigma_i) - \Pi_t|$
\STATE \textbf{Output:} The best solution $\sigma_i^*$
\vspace{0.5em}

\STATE Initialize $n$ fireworks randomly 
\FOR{$t = 1$ to $T$}
    \STATE \textbf{Explode:} Generate $m$ sparks $s$ for each firework $\sigma_i$
    \STATE Initialize a random $bias$ and compute the sparks:
    \[
    s = \sigma_i + A \times bias
    \]
    \STATE Evaluate the fitness $f(s)$ for all sparks $s$ and store both $f(s)$ and $s$ in the SPK RAM
    \STATE Select the elite sparks and store them in the POP RAM
    \STATE \textbf{Mutate:} Compute the fitness difference of sparks:
    \[
    \delta = f(s)_{\text{max}} - f(s)_{\text{min}}
    \]
    \STATE Generate mutated fireworks:
    \[
    \sigma_i^{mut} = \sigma_i + \delta
    \]
    \STATE Clip $\sigma_i^{mut}$ to the search space boundaries $[\sigma_l, \sigma_h]$ and store the fitness $f(\sigma_i^{mut})$ and $\sigma_i^{mut}$ in the MUT RAM
    \STATE \textbf{Select:} Combine all sparks, fireworks, and mutated fireworks, and select the top $n$ solutions based on their fitness values as fireworks in next generation 
    
\ENDFOR
\STATE Select the best solution $\sigma_i^*$ from the $n$ solutions in the last generation
\STATE \textbf{Return:} The best solution $\sigma_i^*$
\end{algorithmic}
\end{algorithm}

\subsubsection{Proposed Custom Dataflow Acceleration Circuit} 
\ding{182} Challenge of Accelerating FWA on Edge. 
The FWA belongs to the class of evolutionary algorithms, requiring iterative searches to find the optimal solution. However, the substantial overhead associated with iterative searches presents a challenge due to the limited energy constraints of edge devices. The design of the FWA accelerator circuit achieves a balance between the algorithm's search efficiency and the hardware overhead. 
\ding{183} Custom Dataflow Circuit.
The dataflow of this circuit is aligned with that of the fireworks algorithm. We have explored the dataflow of FWA and designed a custom accelerator specifically for FWA, avoiding the instruction-fetching and decoding processes typically found in general-purpose CPU/GPU, thereby improving computational efficiency.

\begin{figure}[!t]
\centering
\includegraphics[width=1\linewidth]{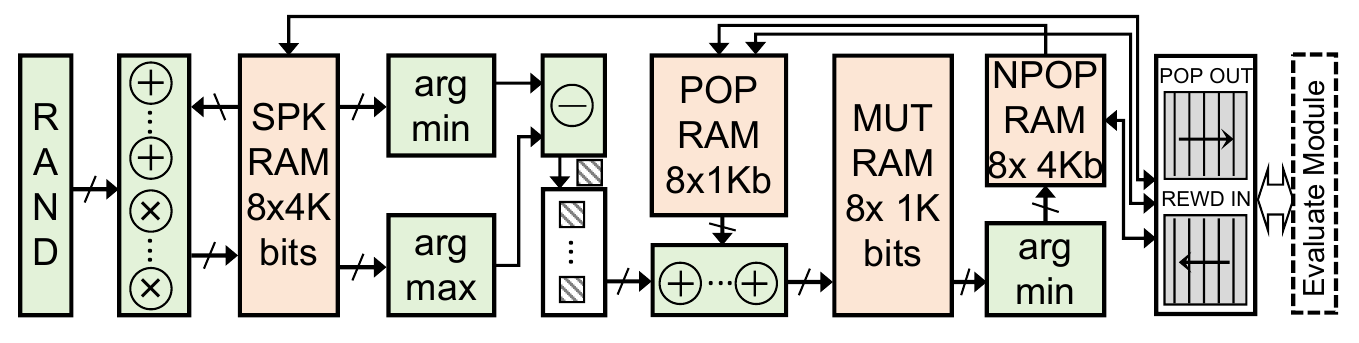}
\caption{The accelerator of FWA-Optimized Search.}
\label{hardware_FWA}
\end{figure}

In the hardware acceleration implementation of the FWA, as shown in Fig.~\ref{hardware_FWA} the system adopts a modular design that covers key functionalities including random number generation, population storage and operations, mutation operations, and fitness evaluation, aiming to optimize hardware resource utilization.

First, the random number generation module (RAND) provides the necessary random values for subsequent modules, supporting the initialization and perturbation of individual and spark positions. During the population storage phase, the SPK RAM (8 × 4K bits) stores the current population's spark information. After performing arithmetic operations such as addition and multiplication on the input random numbers, the generated spark individuals are written into the SPK RAM. An arg min unit is then used to select the spark with the best fitness. The elite individual is stored in the POP RAM (8 × 1 Kb). Meanwhile, the MUT RAM (8 × 1K bits) module performs mutation operations on the population to enhance individual diversity and promote global search capabilities. Mutation results are recombined through addition accumulators and logic operators to generate a new group of individuals. The new spark individuals are subsequently stored in the NPOP RAM (8 × 4Kb), where the system again performs fitness evaluations and selects the new optimal solution using the arg min unit. 


Additionally, the hardware design includes an Evaluate Module, which is responsible for computing and evaluating the fitness of individuals stored in the NPOP RAM. The output results are transmitted via the POP OUT port, while the REWD IN interface supports a feedback mechanism, allowing the optimal individual to be reintroduced into the system, forming an iterative loop.

In the overall circuit design, data transmission between the RAM modules and computational units is implemented in a pipelined manner, ensuring efficient dataflow and parallel computation, which further accelerates the convergence of FWA.

\begin{figure}[!t]
\centering
\includegraphics[width=1\linewidth]{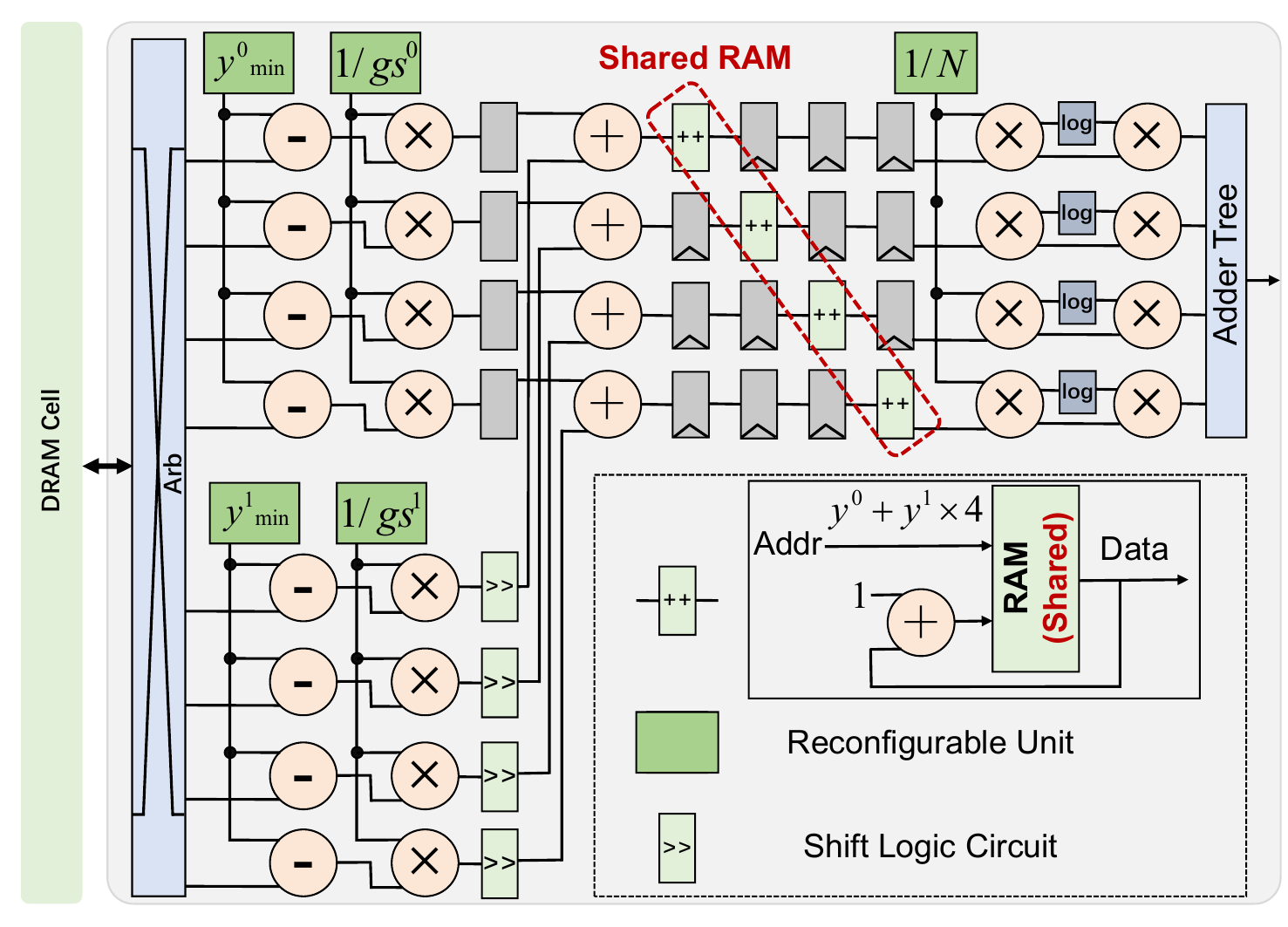}
\caption{Grid-based entropy calculation accelerator. }
\label{grid_based_entropy}
\end{figure}

\subsection{Entropy-Guided Optimization and Hardware}
\subsubsection{Challenge of Detecting Uneven Distribution} Evaluating the quality of the distribution after dimensionality reduction for the traditional algorithms (\eg, Trustworthiness, Continuity and KL divergence) remains highly challenging. Specifically, Trustworthiness method\cite{trustworthiness} measures whether the neighbors of a point in the low-dimensional space are also its neighbors in the high-dimensional space, focusing on false positive neighbors introduced during embedding. However, this method cannot effectively detect issues with uneven distribution, such as outliers,which often lack close neighbors in the high-dimensional space, making Trustworthiness ineffective at distinguishing legitimate outliers from embedding artifacts. Similarly, Continuity\cite{trustworthiness-continuity} algorithm evaluates whether neighbors in the high-dimensional space are preserved in the low-dimensional space, emphasizing false negatives, but it also struggles to detect incorrectly embedded outliers due to the inherent sparsity of their high-dimensional neighborhoods.

In addition, both Trustworthiness and Continuity require the predefinition of a neighborhood size, which limits their generalizability across different datasets. Moreover, KL divergence\cite{tsne_hiton}, which measures the overall similarity between the high-dimensional probability distribution $P_{ij}$ and the low-dimensional probability distribution $Q_{ij}$ , lacks direct interpretability for ranking embedding quality. In low-dimensional embeddings, outliers typically correspond to very small $P_{ij}$ and $Q_{ij}$ values, indicating weak relationships with other points. Consequently, mismatches involving small $P_{ij}$ and $Q_{ij}$ contribute negligibly to the overall KL divergence. As a result, KL divergence performs poorly in penalizing outlier embedding errors.

\subsubsection{Intuition of Entropy} Entropy-based evaluation methods offer unique advantages. As a measure of system uncertainty and information content, entropy can effectively quantify the uniformity of data distributions. During the dimensionality reduction process, entropy serves as an indicator of how uniformly data points are distributed in the low-dimensional space. A higher entropy value reflects a more uniform distribution of data points in low-dimensional space, implying reduced information loss during dimensionality reduction. Conversely, a lower entropy value may indicate clustering or potential information degradation.

Specifically, the advantages of entropy are reflected in the following aspects:
    \ding{182} No need for predefined neighborhoods: Entropy-based methods compute the overall distribution probabilities without requiring a predefined neighborhood size, unlike Trustworthiness and Continuity, thereby offering better generalization across different datasets.
    \ding{183} High sensitivity: Entropy is highly sensitive to changes in data distribution, particularly in detecting outliers and distribution abnormalities. Previous studies \cite{yuan2021fuzzy,xu2022emerging,shah2023entropy} have shown that entropy-based detection methods can effectively identify anomalous patterns in datasets and improve detection rates.
    \ding{184} Global perspective: Prior research \cite{rajendran2022heart,li2023entropy,deveci2022interval} indicates that entropy evaluates the uniformity of data distribution from a global perspective, avoiding the biases that may arise from focusing solely on local neighborhoods.

\subsubsection{Proposed Method}Based on the aforementioned advantages, we propose grid-based entropy as a feedback signal to detect uneven sample distributions and outliers, along with a dedicated accelerator for efficient entropy computation, as shown in Fig.~\ref{grid_based_entropy}. The algorithm is described as follows:

\begin{figure}[!t]
\centering
\includegraphics[width=0.75\linewidth]{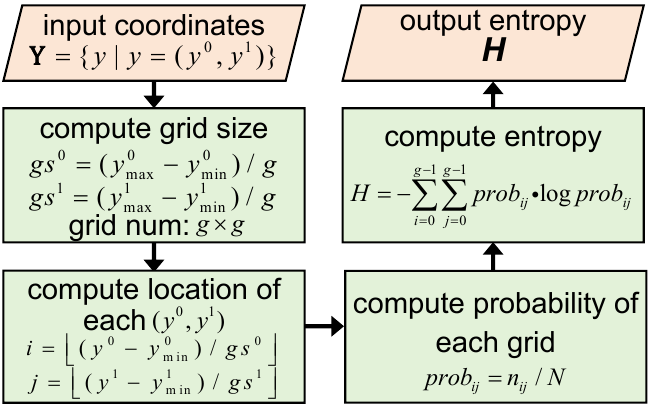}
\caption{Grid-based Entropy Calculation Flowchart. }
\label{entropy_calculation_flowchart}
\end{figure}

\subsubsection{Proposed Time-Multiplexed Acceleration Circuit}
\ding{182} Challenge of Accelerating Entropy-Guided optimization on Edge.
Due to the large number of sample classes, expanding the entropy-guided calculation would result in significant area overhead. Therefore, a reconfigurable time-multiplexed accelerator circuit is designed to achieve higher circuit efficiency and better address the challenge of limited resources in edge devices.
\ding{183} Time-Multiplexed Circuit.
We analyzed the reconfigurability of the variables in the grid-based entropy calculation and designed a time-multiplexed circuit tailored to the algorithm's computational pattern.

In the accelerator implementation of the entropy-guided calculation, as shown in Fig. \ref{grid_based_entropy}, the circuit combines multi-way parallelism and time-division multiplexing. specifically, The circuit employs a four-way parallel computation to enhance data concurrency. First, for low-dimensional space points $\{y|y=(y^0,y^1)\}$, four coordinates are processed simultaneously in each clock cycle. The position in the grid of each coordinate is then calculated, and the number of data points within each grid is updated in the shared RAM to provide data support for subsequent entropy calculations. It is worth noting that the grid base address $(y^0_{min},y^1_{min})$ and grid size $gs^0 \times gs^1$ are reconfigurable, while the shared RAM is time-multiplexed. In the following stage, as shown in Fig.~\ref{entropy_calculation_flowchart}, the sample probability distribution is estimated using frequency-based probability estimation, followed by the computation of the entropy of the distribution to assess whether the low-dimensional space distribution is even.


\section{Experiments and Analysis}
\subsection{Experimental Setups}

\textbf{Dataset:} To evaluate the training accuracy of our proposed algorithm, we conducted experiments on image datasets. 
Based on the experimental setups described in the literature \cite{DQ}, \cite{prakriya2023nessa}, \cite{DQAS}, we selected six datasets: CIFAR10, CINIC10, SVHN, CIFAR100, ImageNet-100, and ImageNet-1K. To assess the performance of the sampling training algorithm on datasets of varying scales, we divided the datasets into small-scale and large-scale groups. The small-scale datasets, which include CIFAR10, CINIC10, and SVHN, consist of 10 classes with images of size 32×32×3. The large-scale datasets consist of CIFAR100, ImageNet-100, and ImageNet-1K. CIFAR100 and ImageNet-100 contain 100 classes, with image dimensions of 32×32×3 and 224×224×3, respectively, while ImageNet-1K includes 1000 classes with images of size 224×224×3. In our comparison with state-of-the-art methods, including DQ \cite{DQ}, DQAS \cite{DQAS}, and NeSSA \cite{prakriya2023nessa}, we found that the DQ and DQAS methods are only implemented for CIFAR10. Therefore, we replicated the DQ and DQAS experiments on the other datasets to ensure a thorough comparison.

In terms of LLM fine-tuning datasets, following \cite{DQ}, we selected the Alpaca dataset as the fine-tuning dataset, which contains approximately 52K instruction-response pairs. The Alpaca dataset is generated using the Self-Instruct method\cite{self-instruct}. To evaluate the performance of the fine-tuned LLMs, we followed the protocol in\cite{instructeval} and selected InstructEval as the evaluation dataset. InstructEval is a comprehensive benchmark specifically designed for assessing instruction-tuned LLMs, with a focus on evaluating their capabilities in problem solving, writing skills, and alignment with human values.

\textbf{DNN Models:} Following prior works \cite{zhao2023dataset}, \cite{DQ}, \cite{prakriya2023nessa}, \cite{DQAS}, we conducted experiments using the PyTorch framework to evaluate a range of DNN and Transformer models, including ResNet-18 \cite{he2016deep}, ResNet-50 \cite{he2016deep}, ShuffleNetV2 \cite{ma2018shufflenet}, MobileNetV2 \cite{sandler2018mobilenetv2}, and ViT \cite{dosovitskiy2020image}. We trained each model for 200 epochs on CIFAR-10 using a batch size of 64 and a cosine annealed learning rate initialized at 0.1. The sample retention rates were set to 10\%, 20\%, and 30\%, respectively. For LLM instruction fine-tuning, following previous studies \cite{DQ}, we selected the LLaMA-7B model and applied LoRA (Low-Rank Adaptation of LLMs) \cite{hu2022lora} for fine-tuning.

\textbf{Hardware:} The hardware server used in our experiments is equipped with an AMD $EPYC^{TM}$ 7H12 64-Core Processor, four NVIDIA GeForce RTX A6000 Ada GPUs, and 512 GB of system memory. To assess the performance of our AdapSNE hardware acceleration architecture, we developed a cycle-accurate simulator to measure the system's latency and memory access patterns. The hardware architecture is implemented in Verilog RTL and synthesized using Synopsis Design Compiler with TSMC 28 nm technology. 

\subsection{Small-Scale Datasets}
 As shown in TABLE \ref{tab:small_scale_dataset}, across the CIFAR10 dataset, AdapSNE achieves an average improvement of 3.9\% in Top-1 accuracy over DQ, 3.8\% over DQAS, and 2.8\% over NeSSA. Compared to NMS, the average accuracy gain is 1.0\%. The maximum observed improvements are 10.2\%, 9.3\%, 4.9\%, and 1.9\%, respectively. On the CINIC10 dataset, AdapSNE demonstrates average accuracy gains of 4.6\% over DQ, 5.2\% over DQAS, and 3.6\% over NeSSA. Relative to NMS, the average improvement is 1.0\%. The maximum accuracy improvements are 16.6\%, 15.1\%, 8.3\%, and 2.0\%, respectively. Across the SVHN dataset, AdapSNE achieves an average improvement of 4.0\% in Top-1 accuracy over DQ, 4.4\% over DQAS, and 1.9\% over NeSSA. Compared to NMS, the average accuracy gain is 0.6\%. The maximum observed improvements are 14.7\%, 13.0\%, 4.8\%, and 1.5\%, respectively.

\begin{table}[ht]
    \setlength\tabcolsep{2.0pt}   
    \centering  
    \caption{The Top-1 accuracy comparison with SOTA dataset compression methods.}		
    \label{tab:small_scale_dataset}
    \begin{tabular}{@{}l|c|c|ccccc|c@{}}
    \toprule
Dataset & Method & KR(\%) & R18 & R50 & ShfV2 & MbV2 & ViT & Avg.(\%) \\
\cline{2-9}
\multirow{16}{*}{CIFAR10}
& Full data & 100 & 95.6 & 95.5 & 85.0 & 90.2 & 80.2 & 89.3 \\
\cline{2-9}
& \multirow{3}{*}{DQ\cite{DQ}}
& 10 & 85.2 & 81.4 & 73.4 & 74.5 & 52.6 & 73.4 \\
& & 20 & 89.4 & 84.9 & 81.6 & 81.4 & 65.9 & 80.6 \\
& & 30 & 91.8 & 89.9 & 81.8 & 86.0 & 71.3 & 84.2 \\
\cline{2-9}
& \multirow{3}{*}{DQAS\cite{DQAS}}
& 10 & 86.1 & 80.8 & 72.3 & 75.7 & 53.5 & 73.7 \\
& & 20 & 90.2 & 85.1 & 80.5 & 80.1 & 66.8 & 80.5 \\
& & 30 & 93.3 & 87.8 & 81.2 & 87.2 & 72.2 & 84.3 \\
\cline{2-9}
& \multirow{3}{*}{NeSSA\cite{prakriya2023nessa}}
& 10 & 87.8 & 81.3 & 72.5 & 76.8 & 58.1 & 75.3 \\
& & 20 & 88.7 & 85.5 & 80.7 & 84.4 & 69.8 & 81.8 \\
& & 30 & 90.5 & 88.2 & 82.8 & 87.4 & 73.6 & 84.5 \\
\cline{2-9}
& \multirow{3}{*}{NMS\cite{nms}}
& 10 & 88.0 & 82.4 & 75.7 & 80.4 & 61.9 & 77.7 \\
& & 20 & 90.4 & 85.7 & 82.4 & 86.4 & 70.3 & 83.0 \\
& & 30 & 93.5 & 90.0 & 84.0 & 88.3 & 75.6 & 86.3 \\
\cline{2-9}
& \multirow{3}{*}{AdapSNE(Ours)}
& 10 & \textbf{88.7} & \textbf{83.1} & \textbf{76.0} & \textbf{81.4} & \textbf{62.8} & \textbf{78.4} \\
& & 20 & \textbf{90.8} & \textbf{85.8} & \textbf{83.7} & \textbf{88.2} & \textbf{70.8} & \textbf{83.9} \\
& & 30 & \textbf{95.4} & \textbf{91.1} & \textbf{85.0} & \textbf{89.3} & \textbf{76.1} & \textbf{87.4} \\
\hline

\multirow{13}{*}{CINIC10}
& Full data & 100 & 82.5 & 81.7 & 74.7 & 78.7 & 76.8 & 78.9 \\
\cline{2-9}
& \multirow{3}{*}{DQ\cite{DQ}}
& 10 & 68.9 & 64.1 & 61.5 & 62.8 & 37.4 & 58.9 \\
& & 20 & 75.9 & 76.6 & 67.8 & 70.9 & 49.0 & 68.0 \\
& & 30 & 79.8 & 80.3 & 70.0 & 75.1 & 55.3 & 72.1 \\
\cline{2-9}
& \multirow{3}{*}{DQAS\cite{DQAS}}
& 10 & 68.0 & 63.1 & 60.2 & 64.1 & 37.6 & 58.6 \\
& & 20 & 72.9 & 76.2 & 66.7 & 69.8 & 50.5 & 67.2 \\
& & 30 & 77.0 & 77.9 & 69.1 & 76.4 & 56.8 & 71.4 \\
\cline{2-9}
& \multirow{3}{*}{NeSSA\cite{prakriya2023nessa}}
& 10 & 69.2 & 63.8 & 61.4 & 65.8 & 40.1 & 60.1 \\
& & 20 & 75.7 & 73.6 & 67.8 & 70.5 & 51.3 & 67.8 \\
& & 30 & 80.3 & 78.8 & 70.8 & 76.8 & 63.6 & 74.1 \\
\cline{2-9}
& \multirow{3}{*}{NMS\cite{nms}}
& 10 & 70.8 & 65.1 & 63.5 & 69.2 & 46.4 & 63.0 \\
& & 20 & 77.2 & 77.0 & 68.5 & 76.3 & 53.3 & 70.5 \\
& & 30 & 81.1 & 80.4 & 72.4 & 77.5 & 69.9 & 76.3 \\
\cline{2-9}
& \multirow{3}{*}{AdapSNE(Ours)}
& 10 & \textbf{72.0} & \textbf{65.6} & \textbf{64.8} & \textbf{69.3} & \textbf{47.1} & \textbf{63.8} \\
& & 20 & \textbf{78.4} & \textbf{77.8} & \textbf{69.9} & \textbf{76.3} & \textbf{54.3} & \textbf{71.3} \\
& & 30 & \textbf{82.5} & \textbf{81.5} & \textbf{73.3} & \textbf{78.1} & \textbf{71.9} & \textbf{77.5} \\
\hline

\multirow{13}{*}{SVHN}
& Full data & 100 & 95.9 & 96.1 & 95.2 & 95.7 & 52.6 & 87.1 \\
\cline{2-9}
& \multirow{3}{*}{DQ\cite{DQ}}
& 10 & 92.5 & 89.1 & 90.3 & 87.5 & 10.2 & 73.9 \\
& & 20 & 93.6 & 92.1 & 93.0 & 89.2 & 26.4 & 78.9 \\
& & 30 & 94.8 & 94.8 & 93.4 & 93.6 & 27.4 & 80.8 \\
\cline{2-9}
& \multirow{3}{*}{DQAS\cite{DQAS}}
& 10 & 92.0 & 89.4 & 89.6 & 89.4 & 11.0 & 74.3 \\
& & 20 & 90.8 & 91.4 & 92.5 & 88.4 & 27.9 & 78.2 \\
& & 30 & 92.6 & 92.4 & 91.5 & 94.7 & 29.1 & 80.1 \\
\cline{2-9}
& \multirow{3}{*}{NeSSA\cite{prakriya2023nessa}}
& 10 & 92.9 & 90.1 & 90.5 & 89.6 & 19.5 & 76.5 \\
& & 20 & 95.2 & 91.9 & 92.8 & 90.7 & 29.8 & 80.1 \\
& & 30 & 95.3 & 93.4 & 92.9 & 94.6 & 40.8 & 83.4 \\
\cline{2-9} 
& \multirow{3}{*}{NMS\cite{nms}}
& 10 & 93.8 & 90.4 & 92.7 & 93.6 & 19.6 & 78.0 \\
& & 20 & 95.4 & 92.3 & 94.1 & 94.6 & 30.8 & 81.4 \\
& & 30 & 95.6 & 95.1 & 94.9 & 95.2 & 41.2 & 84.4 \\
\cline{2-9}
& \multirow{3}{*}{AdapSNE(Ours)}
& 10 & \textbf{94.6} & \textbf{90.9} & \textbf{93.2} & \textbf{94.4} & \textbf{20.0} & \textbf{78.6} \\
& & 20 & \textbf{95.5} & \textbf{93.8} & \textbf{94.7} & \textbf{95.2} & \textbf{31.7} & \textbf{82.2} \\
& & 30 & \textbf{95.8} & \textbf{95.4} & \textbf{94.9} & \textbf{95.4} & \textbf{42.1} & \textbf{84.7} \\
\bottomrule
\end{tabular}
\label{exp_vs_dataselection}
\end{table}

\subsection{Large-Scale Datasets}
\begin{figure*}[!t]
\centering
\includegraphics[width=0.75\linewidth]{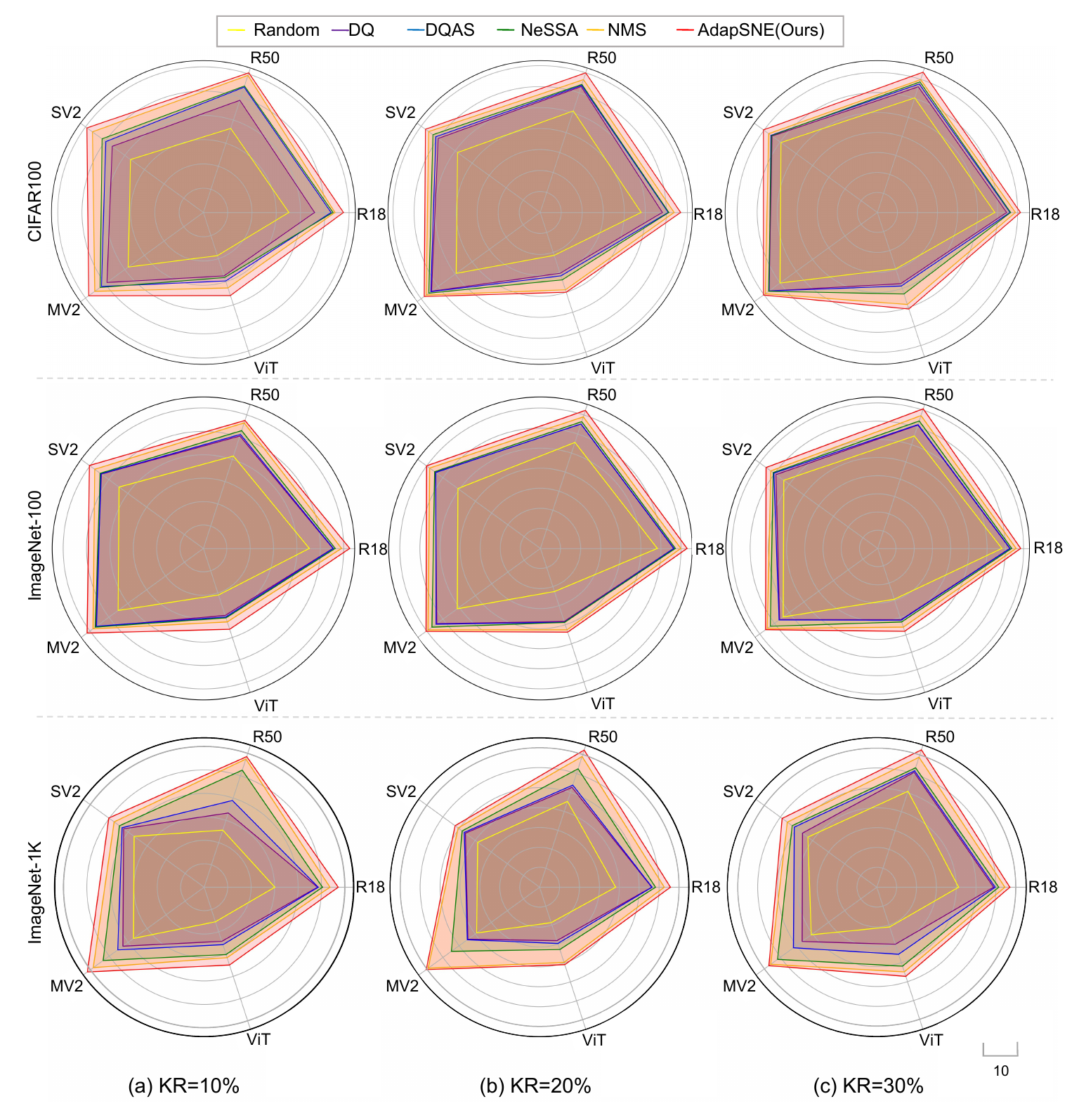}
\caption{Radar chart comparison of Top-1 accuracy across five network architectures (MobileNetV2, ShuffleNetV2, ResNet50, ResNet18, ViT) on ImageNet-1K, ImageNet-100, and CIFAR100, under compression ratios of 10\%, 20\%, and 30\%. AdapSNE outperforms existing methods including NMS, DQ, DQAS, and NeSSA.}
\label{large_scale_dataset}
\end{figure*}

As shown in Fig. \ref{large_scale_dataset}, Across the ImageNet100 dataset, AdapSNE achieves an average improvement of 6.7\% in Top-1 accuracy over DQ, 6.4\% over DQAS, and 5.1\% over NeSSA. Compared to NMS, the average accuracy gain is 2.5\%. The maximum observed improvements are 9.5\%, 9.1\%, 7.0\%, and 3.9\%, respectively. On the CIFAR100 dataset, AdapSNE demonstrates average accuracy gains of 8.5\% over DQ, 6.4\% over DQAS, and 5.5\% over NeSSA. Relative to NMS, the average improvement is 2.5\%. The maximum accuracy improvements are 13.3\%, 12.1\%, 7.9\%, and 3.9\%, respectively. Across the ImageNet-1K dataset, AdapSNE achieves an average improvement of 14.4\% in Top-1 accuracy over DQ, 12.2\% over DQAS, and 7.3\% over NeSSA. Compared to NMS, the average accuracy gain is 2.5\%. The maximum observed improvements are 25.9\%, 25.5\%, 15.6\%, and 3.9\%, respectively. 

\subsection{LLM Datasets}

\begin{figure*}[!t]
\centering
\includegraphics[width=1\linewidth]{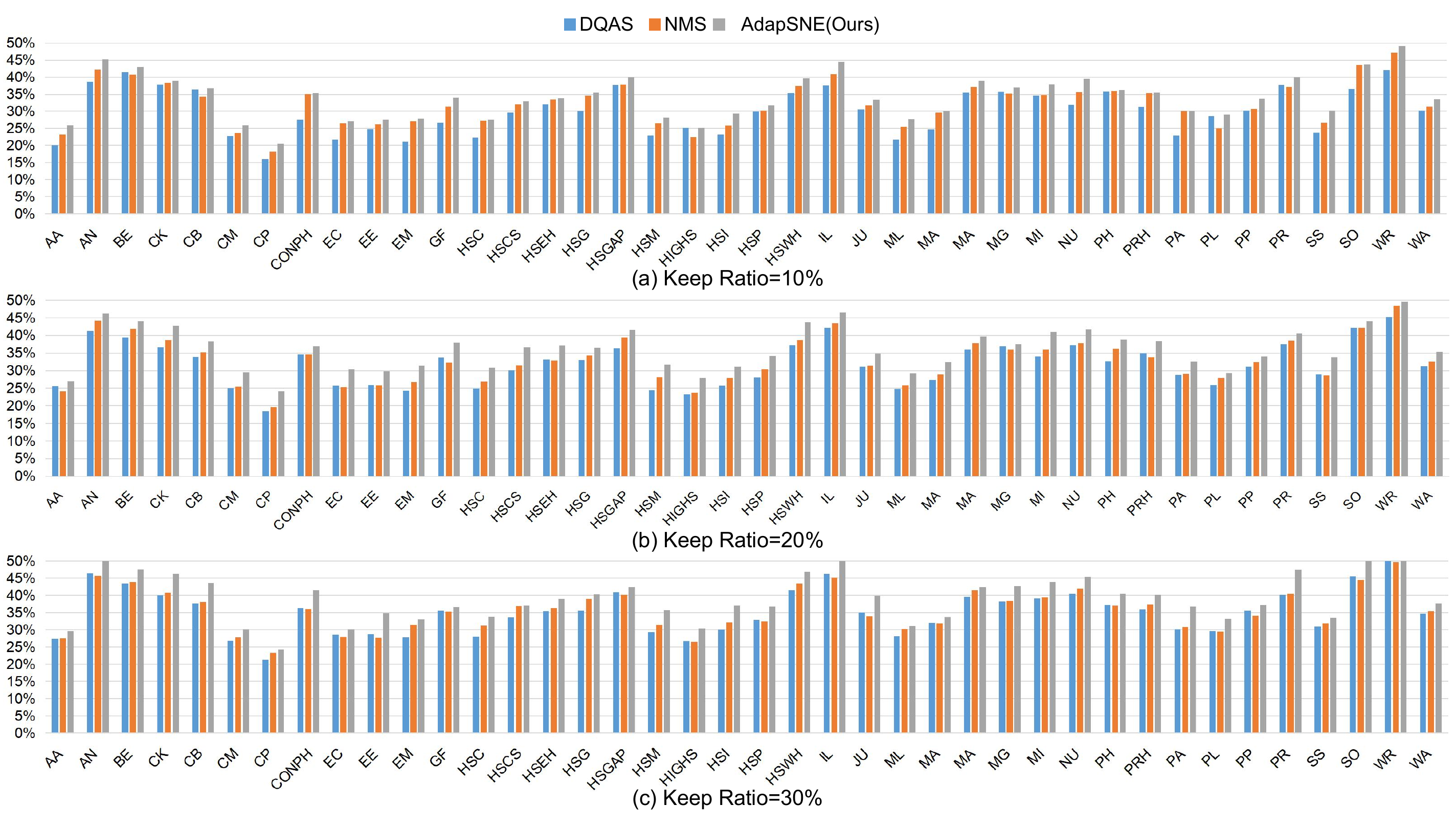}
\caption{Accuracy comparison of the LLM dataset at keeping ratios of 10\% (a), 20\% (b), and 30\% (c). }
\label{LLM_Dataset}
\end{figure*}

As shown in Fig. \ref{LLM_Dataset}. At a keeping ratio of 10.0\%, AdapSNE achieves an average improvement of 3.4\% in Top-1 accuracy over DQSA and 2.2\% over NMS. The maximum observed improvements are 7.7\% and 4.0\%, respectively. At a keeping ratio of 20.0\%, AdapSNE demonstrates an average accuracy gain of 4.1\% over DQSA and 2.8\% over NMS. The maximum accuracy improvements are 7.3\% and 5.7\%, respectively. At a keeping ratio of 30.0\%, AdapSNE achieves an average improvement of 2.9\% in Top-1 accuracy over DQSA and 2.1\% over NMS. The maximum observed improvements are 7.2\% and 7.1\%, respectively. 

\section{EVALUATION}

\subsection{Hardware Overhead Breakdown}


As part of the overhead analysis for the proposed FWA Search and entropy-guided optimization circuits, we report a full power and area breakdown of all hardware blocks. As shown in Fig.~9, the conventional t-SNE pipeline dominates both power and silicon footprint, whereas the added FWA Search and entropy-guided optimization engines incur only modest overhead—2.7\% of total power and 6.1\% of total area. This efficiency is primarily attributed to our custom dataflow and time-multiplexed design, which effectively amortize control and arithmetic resources.

\begin{figure}[t]
\centering
\includegraphics[width=0.95\linewidth]{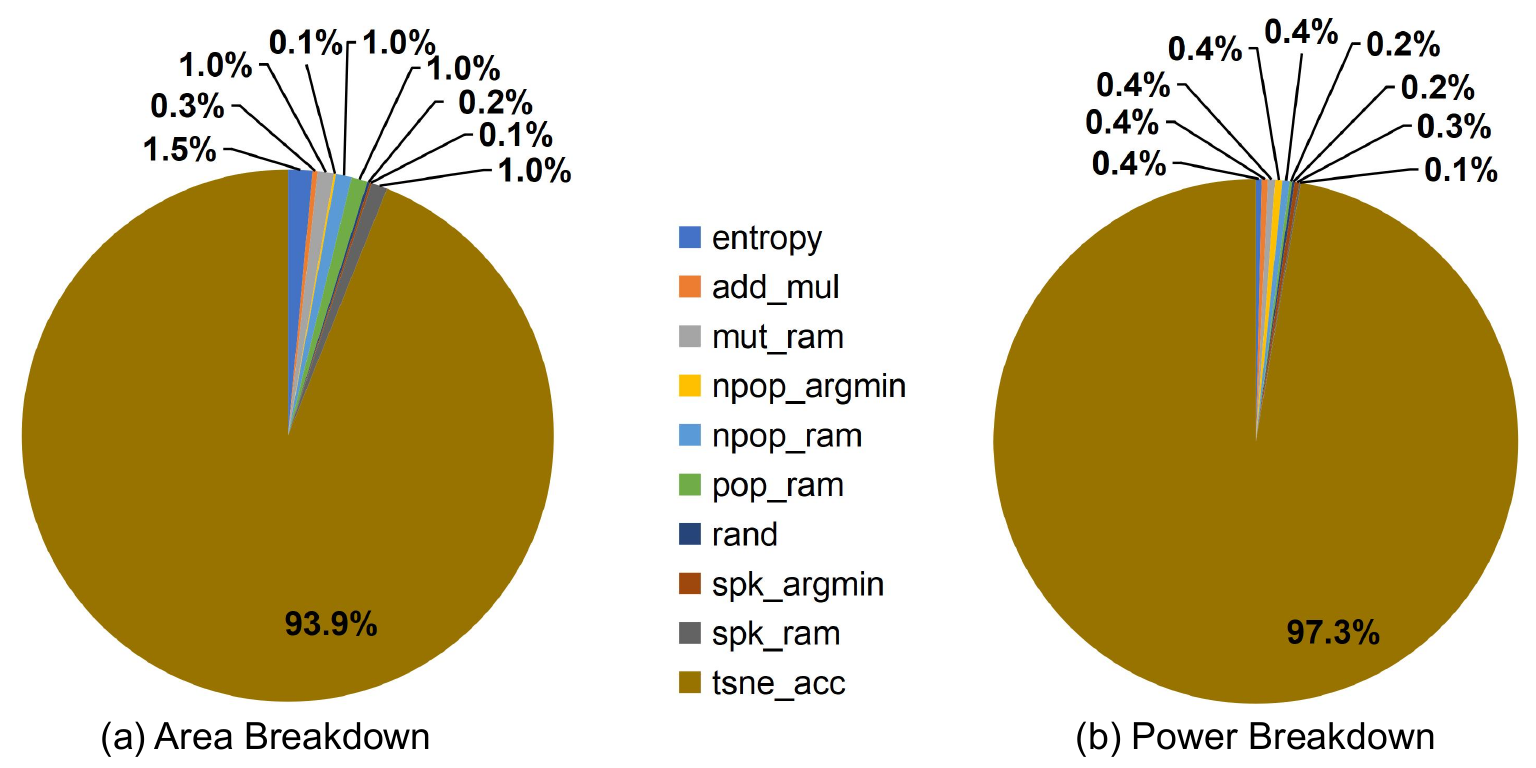}
\caption{(a) Area breakdown of the AdapSNE hardware modules. (b)Power breakdown of the AdapSNE hardware modules.}
\label{area_breakdown}
\end{figure}

\subsection{Ablation Study}

\begin{table}[t]
\centering
\caption{Comparison of t-SNE and AdapSNE on Different Datasets and Keep Ratios. }
\label{tab:adapsne_comparison}
\begin{tabular}{c|c|c|c|c|c}
\toprule
 Dataset           &     Method         & 10\%            & 20\%            & 30\%         &  Avg(\%)   \\ \hline
\multirow{2}{*}{CIFAR10}     & t-SNE        & 82.3          & 87.6          & 90.2         &  86.7    \\ \cline{2-6}
                             & AdapSNE & \textbf{88.7} & \textbf{90.8} & \textbf{95.4}      &  \textbf{91.6}    \\     \hline      
\multirow{2}{*}{CIFAR100}    & t-SNE        & 50.4          & 61.2          & 65.2         &  58.9    \\ \cline{2-6}
                            & AdapSNE & \textbf{57.7} & \textbf{66.8} & \textbf{71.4}       &  \textbf{65.3}    \\ \hline
\multirow{2}{*}{ImageNet-1K} & t-SNE        & 48.7          & 58.2          & 61.0         &  56.0     \\ \cline{2-6}
                            & AdapSNE & \textbf{57.1} & \textbf{65.7} & \textbf{67.0}       &  \textbf{63.3}   \\
\bottomrule
\end{tabular}
\label{vs_tsne}
\end{table}

As shown in TABLE~\ref{tab:adapsne_comparison}, the ablation study demonstrates the effectiveness of AdapSNE, which incorporates the Entropy-guided optimization mechanism and the FWA-optimized search algorithm, by achieving consistent accuracy improvements over the baseline t-SNE method across multiple datasets and keeping ratios. 
The Entropy-guided optimization mechanism iteratively optimizes the target perplexity, and the FWA-optimized search method improves the convergence of the perplexity error function based on the target perplexity. Both work simultaneously, which is why we perform ablation studies considering both factors. On CIFAR10, AdapSNE outperforms t-SNE with an average accuracy gain of 4.9\% (91.6\% vs. 86.7\%), while on CIFAR100, the improvement rises to 6.4\% (65.3\% vs. 58.9\%). Notably, for the large-scale ImageNet-1K dataset, AdapSNE achieves an average gain of 7.3\% (63.3\% vs. 56.0\%), with a maximum improvement of 8.4\% at a KR of 10\%. These results highlight the ability of AdapSNE to deliver superior accuracy by better preserving information and enhancing robustness, particularly on complex and high-dimensional datasets, validating the advantages of the proposed mechanisms.

\section{Conclusion}

In conclusion, this paper addresses critical challenges in training DNNs directly on resource-constrained edge devices, particularly the prohibitive overhead associated with large datasets and the poor generalization caused by \textit{representative bias} in existing DNN-free sample selection methods. To overcome these issues, we introduced AdapSNE, a novel DNN-free sample selection algorithm. AdapSNE innovatively integrates two key techniques: (1) the Fireworks Algorithm to efficiently handle the non-monotonic perplexity error and suppress outliers in the reduced representation space; (2) entropy-guided optimization to ensure uniform sampling. By tackling the two core limitations of the prior state-of-the-art method, AdapSNE generates significantly more representative training samples, leading to enhanced training accuracy. Furthermore, to minimize the computational overhead of AdapSNE's iterative FWA and optimization steps, we designed a specialized hardware accelerator featuring custom dataflow and time-multiplexing techniques, achieving dramatic reductions in on-device training energy consumption and silicon area. Comprehensive experimental evaluations convincingly demonstrate the generalizability and effectiveness of AdapSNE.


\bibliographystyle{IEEEtran}



\end{document}